\documentclass[sigconf]{acmart}

\setcopyright{acmcopyright}
\renewcommand\footnotetextcopyrightpermission[1]{}

\AtBeginDocument{%
  \providecommand\BibTeX{{%
    \normalfont B\kern-0.5em{\scshape i\kern-0.25em b}\kern-0.8em\TeX}}}

\usepackage[flushleft]{threeparttable}
\usepackage{subfigure}
\usepackage{xspace}

\newcommand{\para }[1]{\emph {\textbf {#1}}}

\copyrightyear{2023} 
\acmYear{2023} 
\setcopyright{acmlicensed}\acmConference[HotCarbon '23]{2nd Workshop on Sustainable Computer Systems}{July 9, 2023}{Boston, MA, USA}
\acmBooktitle{2nd Workshop on Sustainable Computer Systems (HotCarbon '23), July 9, 2023, Boston, MA, USA}
\acmPrice{15.00}
\acmDOI{10.1145/3604930.3605708}
\acmISBN{979-8-4007-0242-6/23/07}

\usepackage{amsmath,amsfonts,bm}

\def\eqref#1{equation~\ref{#1}}

\def\1{\bm{1}}

\def\rr{{\textnormal{r}}}

\def\vf{{\bm{f}}}

\def\vx{{\bm{x}}}
\def\vy{{\bm{y}}}

\DeclareMathAlphabet{\mathsfit}{\encodingdefault}{\sfdefault}{m}{sl}
\SetMathAlphabet{\mathsfit}{bold}{\encodingdefault}{\sfdefault}{bx}{n}

\newcommand{\sys}{\texttt{CE-NAS}\xspace}

\begin{document}

\title{Carbon-Efficient Neural Architecture Search}

\author{Yiyang Zhao}
\affiliation{%
  \institution{Worcester Polytechnic Institute}  
  \country{}}
\email{yzhao10@wpi.edu}

\author{Tian Guo}
\affiliation{%
  \institution{Worcester Polytechnic Institute}
  \country{}
  }
\email{tian@wpi.edu}

\begin{abstract}
  This work presents a novel approach to neural architecture search (NAS) that aims to reduce energy costs and increase carbon efficiency during the model design process. The proposed framework, called carbon-efficient NAS (\sys), consists of NAS evaluation algorithms with different energy requirements, a multi-objective optimizer, and a heuristic GPU allocation strategy. \sys dynamically balances energy-efficient sampling and energy-consuming evaluation tasks based on current carbon emissions. Using a recent NAS benchmark dataset and two carbon traces, our trace-driven simulations demonstrate that \sys achieves better carbon and search efficiency than the three baselines. 
\end{abstract}



\maketitle
\pagestyle{plain}
\pagenumbering{gobble}

\section{Introduction}
\label{sec:intro}
Deep Learning (DL) has become an increasingly important field in computer science, with applications ranging from healthcare to transportation to energy management. However, DL training is notoriously energy-intensive and significantly contributes to today's carbon emissions~\cite{energy_cost, nasnet}. The main culprit comes down to the iterative nature of training, which requires evaluating and updating model parameters based on a large amount of data. 

Neural architecture search (NAS) has emerged as a means to automate the design of DL models. At the high level, NAS often involves leveraging search algorithms to explore a massive architecture design space, ranging from hundreds of millions to trillions of candidates~\cite{nasnet, rea, alphax, few-shot, lamoo, DARTS}, by training and evaluating a subset of architectures. In searching for the best architecture for different application domains, many NAS works have reported using thousands of GPU-hours~\cite{nasnet, rea, alphax, lanas, lamoo}.

The environmental impact of NAS, if left untamed, can be substantial. While recent works have significantly improved the search efficiency of NAS~\cite{nasnet, rea, alphax, lanas,few-shot}, e.g., reducing the GPU-hours to tens of hours without sacrificing the architecture quality, there still lacks \emph{conscious efforts in reducing carbon emissions}. As noted in a recent vision paper by Bashir et al.~\cite{carbon_time}, energy efficiency can help reduce carbon emissions but is not equivalent to carbon efficiency. This paper aims to bridge the gap between carbon and energy efficiency with a new NAS framework designed to be carbon-aware from the outset.   

The proposed framework, termed \sys, will tackle the high carbon emission problem from two main aspects. First, \sys will regulate the model design process by \emph{deciding when to use different NAS evaluation strategies based on the varying carbon intensity}. To elaborate, given a fixed amount of GPU resources, \sys will allocate more resources to energy-efficient NAS evaluation strategies, e.g., one-shot NAS ~\cite{enas, BG_understanding, DARTS, few-shot, proxylessnas, ofa}, during periods of high carbon intensity. Conversely, during periods of low carbon intensity, \sys will shift the focus to running energy-intensive but more effective NAS evaluation strategies, e.g., vanilla NAS~\cite{nasnet, rea, alphax, lanas}.
Second, the \sys framework will support energy and carbon-efficient DL model design via multi-objective optimization. Specifically, we will leverage a recent learning-based multi-objective optimizer LaMOO~\cite{lamoo} and integrate it to \sys to achieve search efficiency.

\begin{figure*}[t]
\centering
  \includegraphics[width=0.96\textwidth]{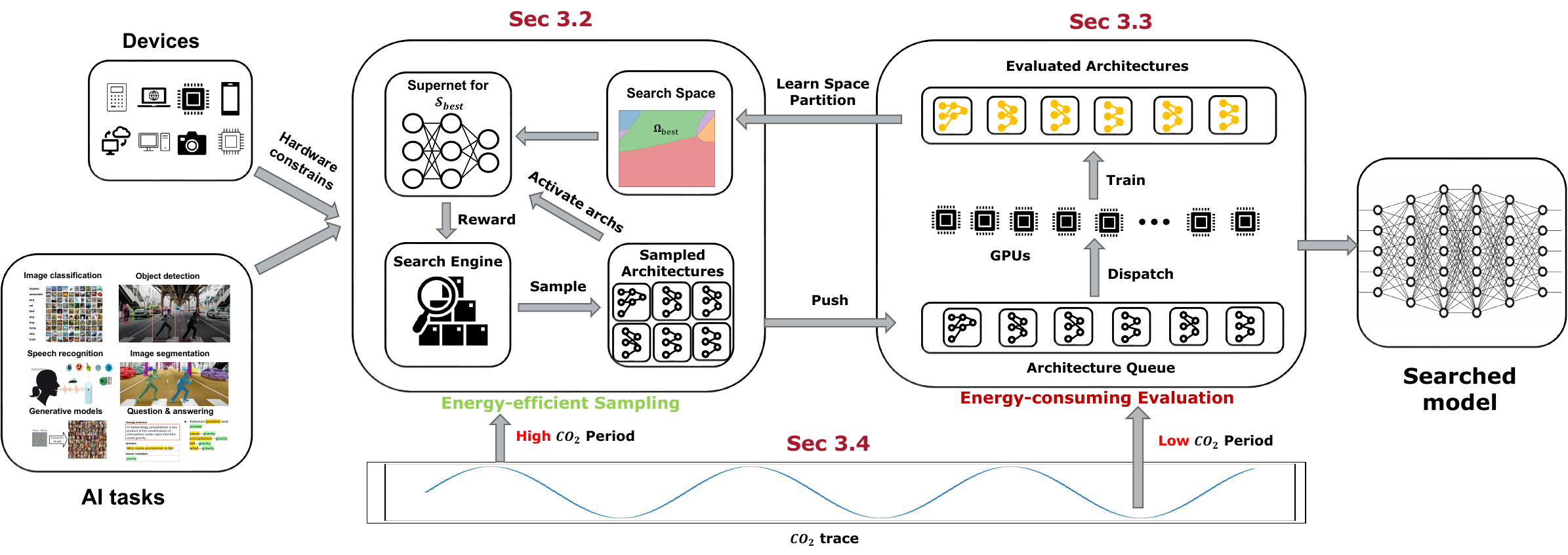}
  \caption{An overview of \sys. 
  \textnormal{The sampling and evaluation tasks will be dispatched with different GPU resources based on carbon emission intensity during the neural architecture search.
  }
  }
  \label{fig:solution_overview}
\end{figure*}

Based on these two design guidelines, we sketch out the basis of the proposed \sys framework in Figure~\ref{fig:solution_overview} and implement a trace-driven simulator to investigate the promise of \sys in improving carbon and search efficiencies. Using carbon emission traces from electricityMap~\cite{electricitymap} and a new NAS dataset called HW-NASBench~\cite{hwnasbench}, we show that \sys has the least relative carbon emissions and only marginally lower search efficiency compared to vanilla LaMOO~\cite{lamoo}. Based on our investigation, we believe there are many fruitful directions in the context of \sys which we outline in \S\ref{sec:conclusion}. We hope this discussion will serve as the blueprint and a baseline for building a carbon-efficient NAS framework.  

\begin{table}[t]
\centering
  \caption{Comparison of energy-efficient NAS evaluation methods. \textnormal{ \emph{Eval. cost} refers to the cost of obtaining the evaluation results. \emph{Init. cost} describes additional dataset preparation and the time required for training the model (e.g., supernet or predictor). \emph{Accuracy} measures the rank correlation between the evaluation method and the actual rank. Predictor-based methods require \emph{Extra data} as a training set to construct the prediction model.}}
  \label{tab:energy_aware_nas_table}
  \resizebox{0.45\textwidth}{!}{
  \begin{threeparttable}
\begin{tabular}{@{}lcccc@{}}
\toprule
\multicolumn{1}{c}{\textbf{Method}}  & \textbf{Eval. cost}  & \textbf{Init. cost} & \textbf{Accuracy}                 & \textbf{Extra data}      \\ \midrule
\multicolumn{1}{|l|}{One-shot~\cite{DARTS, enas, few-shot, pcdarts, ofa}}       & \multicolumn{1}{c|}{Low}  & \multicolumn{1}{c|}{Low}     & \multicolumn{1}{c|}{Intermediate} & \multicolumn{1}{c|}{No}  \\ \midrule
\multicolumn{1}{|l|}{Predictor~\cite{pnas, learning_curve, alphax}}      & \multicolumn{1}{c|}{Low}  & \multicolumn{1}{c|}{High\tnote{$\dagger$}}    & \multicolumn{1}{c|}{High\tnote{$\dagger$}}         & \multicolumn{1}{c|}{Yes} \\ \midrule
\multicolumn{1}{|l|}{Low-fidelity~\cite{nasnet, rea, DARTS, alphax, sub_dataset}}   & \multicolumn{1}{c|}{High} & \multicolumn{1}{c|}{None}    & \multicolumn{1}{c|}{Intermediate\tnote{$\ddagger$}} & \multicolumn{1}{c|}{No}  \\ \midrule
\multicolumn{1}{|l|}{Gradient Proxy~\cite{knas}} & \multicolumn{1}{c|}{Low}  & \multicolumn{1}{c|}{Low}     & \multicolumn{1}{c|}{Intermediate} & \multicolumn{1}{c|}{No}  \\ \bottomrule
\end{tabular}
\begin{tablenotes}
    \item[$\dagger$] It depends on the size of extra data.  
    \item[$\ddagger$] It depends on the extent of the fidelity. 
  \end{tablenotes}
\end{threeparttable}
}
\end{table}

\section{NAS and its Carbon Impact}

Neural architecture search (NAS) is a technique for automating the design of neural network architectures. NAS aims to find an optimal network architecture that performs well on a specific task without human intervention. NAS-designed neural architectures achieved state-of-the-art performances on many AI applications, such as image classification, object detection, and image segmentation~\cite{nasnet, nas-fcos, nas-fpn, rea, alphax}.

However, NAS typically requires significant computational resources (e.g., GPUs) to find the optimal architecture, with most of these resources being used for architecture evaluation. For example, Zoph et al.~\cite{nasnet} used 800 GPUs for 28 days, equivalent to 22,400 GPU hours, to obtain the final architectures. Strubell et al.~\cite{energy_cost} found that a single NAS solution can emit as much carbon as five cars during its lifetime. These findings highlight the need for energy and carbon-efficient NAS methods to reduce the environmental impact of AI research.  

Existing works on energy-efficient NAS often focus on improving the evaluation phase, e.g., via weight-sharing~\cite{enas, DARTS, pcdarts, pdarts, ofa}, performance predictor~\cite{pnas, ofa, alphax, nasbench301}, low-fidelity NAS evaluation~\cite{nasnet, rea, alphax, lanas, DARTS, sub_dataset}, and gradient proxy NAS~\cite{knas}.
A comparison of these methods can be found in Table~\ref{tab:energy_aware_nas_table}. 
Weight-sharing leverages the accuracy estimated by a \emph{supernet} as a proxy for the true architecture performance, while gradient proxy NAS uses the gradient as a proxy. These proxy-based methods, although incurring smaller search costs in terms of energy, can have lower search efficiency because their estimated architecture accuracy may have poor rank correlation~\cite{few-shot}. 
Performance predictors provide a more accurate performance prediction than weight-sharing and gradient proxy NAS. Still, their accuracy heavily relies on the volume and quality of the training data, which can be very expensive to create~\cite{nasbench101, nasbench301}. Low-fidelity evaluation still requires training each searched architecture, leading to limited energy savings. 

In short, utilizing existing energy-efficient NAS methods requires careful consideration of the search quality and efficiency trade-offs; however, naively applying these methods may not even lead to energy savings, not to mention lower carbon emissions. In this work, we achieve the goals of search efficiency, search quality, and carbon efficiency by leveraging a generic multi-objective optimizer~\cite{lamoo}, a mix of energy-efficient~\cite{BG_understanding,enas,DARTS,few-shot} and energy-consuming~\cite{nasnet, rea, alphax, lanas} evaluation methods, and a carbon-aware GPU resource allocation strategy.

\section{Research Roadmap}
\label{sec:roadmaps}

In this section, we present an overview of the proposed \sys framework (Figure~\ref{fig:solution_overview}) and sketch the basis for each component. We hope this discussion will serve as the blueprint and a baseline for designing a carbon-efficient NAS framework.  

\subsection{\sys Overview}
 
As observed in~\cite{carbon_time}, grid carbon emissions vary geographically and temporally based on the mix of active generators. Consequently, different carbon emissions arise even when consuming the same electricity at different locations or times. Operating the NAS framework without considering costs across every carbon period will lead to carbon waste when utilizing carbon-consuming but effective NAS methods. Conversely, employing carbon-saving yet sample-inefficient NAS methods may deteriorate search performance.

To address this issue, we propose a carbon-aware adaptive NAS search strategy that balances energy consumption during high-carbon and low-carbon periods. Our strategy decouples the two parts of a NAS search process---evaluation (energy-consuming) and sampling (energy-saving)---and handles them independently across different carbon periods. The basic idea involves leveraging the energy-efficient one-shot NAS~\cite{BG_understanding, enas, DARTS, few-shot} to effectively estimate the accuracy of architectures in the sampling process during periods of high carbon intensity. Meanwhile, we will run the expensive evaluation part primarily during low-carbon periods. In the following sections, we will provide a detailed explanation of the carbon-aware NAS strategy.

\subsection{Search Initialization}
\label{sec:initialization}

Similar to other optimization problems~\cite{qehvi, lamcts, lamoo, rea}, the first step in our proposed carbon-efficient NAS framework involves initializing the search process by randomly selecting several architectures, $\textbf{a}$, from the search space, $\Omega$, and evaluating their accuracy, $E(\textbf{a})$, carbon emissions, $C(\textbf{a})$, and inference energy, $I(\textbf{a})$. The resulting samples are then added to the observed samples set, $\mathcal{P}$.

Here, we define two types of methods for evaluating the accuracy of architectures. One is actual training, which trains the architecture $a$ from scratch until convergence and evaluates it to obtain its true accuracy, $E(a)$. Another method is called \emph{one-shot evaluation}~\cite{BG_understanding,enas,DARTS}, which leverages a trained \emph{supernet} to estimate the accuracy of the architecture, denoted as $E^{'}(a)$. Note that obtaining $E^{'}(a)$ is energy-efficient; however, due to the co-adaption among operations~\cite{few-shot}, $E^{'}(a)$ is often not as accurate as $E(a)$. We train all the sampled architectures in the initialization stage to obtain their true accuracy for further search.

\subsection{Energy-Efficient Architecture Sampling}
\label{sec:sampling}

To search for architectures with high inference accuracy and low inference energy, we formulate the search problem as a multi-objective optimization (MOO).

\paragraph{Primer.} Mathematically, in multi-objective optimization we optimize $M$ objectives $\vf(\vx) = [f_1(\vx), f_2(\vx), \ldots, f_M(\vx)]\in \rr^M$:
\begin{eqnarray}
\min\;& f_1(\vx), f_2(\vx), ..., f_M(\vx)  \label{eq:problem-setting}  \label{prob-formulation} \\
\mathrm{s.t.}\;& \vx \in \Omega \nonumber, 
\end{eqnarray}

where $f_i(\vx)$ denotes the function of objective $i$. Modern MOO methods aim to find the problem's entire \emph{Pareto frontier}, the set of solutions that are not \emph{dominated} by any other feasible solutions. Here we define \emph{dominance} $\vy \prec_\vf \vx$ as $f_i(\vx) \leq f_i(\vy)$ for all functions $f_i$, and there exists at least one $i$ s.t. $f_i(\vx) < f_i(\vy)$, $1\le i \le M$. If the condition holds, a solution $\vx$ is always better than solution $\vy$.

\paragraph{Multi-objective search space partition.} We leverage the recently proposed multi-objective optimizer called LaMOO~\cite{lamoo} that learns to partition the search space from observed samples to focus on promising regions likely to contain the Pareto frontier. LaMOO is a general optimizer; we can extend it to NAS as follows. 

We utilize LaMOO~\cite{lamoo} to partition the search space, $\Omega$, into several sub-search spaces. This partitioning will be based on the architectures and their true accuracy ($E(\textbf{a})$) and inference energy ($I(\textbf{a})$) as observed in the sample set, $\mathcal{P}$. Specifically, LaMOO recursively divides the search space into promising and non-promising regions. Each partitioned region can then be mapped to a node in a search tree. Using Monte-Carlo Tree Search (MCTS), LaMOO selects the most promising sub-space (i.e., tree node) for further exploration based on their UCB values~\cite{ucb}. This optimal sub-space is denoted as $\Omega_{best}$.

Next, we will construct and train a supernet~\cite{BG_understanding, few-shot}, $\mathcal{S}_{best}$, for $\Omega_{best}$. We then use a NAS search algorithm to identify new architectures that will be used to refine the search space. In this work, we employ the state-of-the-art multi-objective Bayesian optimization algorithm qNEHVI~\cite{qnehvi}. This algorithm will generate new architectures, denoted as $\bm{a_n}$, from $\Omega_{best}$, and estimate their approximate accuracy, $E^{'}(\bm{a_n})$, using $\mathcal{S}_{best}$. At the same time, these architectures $\bm{a_n}$ are added to a ready-to-train set, $\mathcal{R}$, consisting of architecture candidates for further training.

Currently, to avoid unnecessary training and energy consumption, we define the maximum capacity of $\mathcal{R}$ as $Cap(\mathcal{R})$. When the capacity reaches, i.e., when there are more architectures to train than we have resources for, the sampling process blocks until spaces free up in $\mathcal{R}$. The accuracy of architectures, either estimated by $\mathcal{S}_{best}$ or obtained from training, will be fed back into the search engine as shown in Figure~\ref{fig:solution_overview} to repeat the process described above.

As mentioned in \S\ref{sec:initialization}, obtaining estimated accuracy through supernet is energy-efficient because these architectures can be evaluated without the time-consuming training. Therefore, during high carbon emission periods, \sys will try to perform this process to save energy and produce as little carbon as possible, as shown in the middle left part of Figure~\ref{fig:solution_overview}.

\subsection{Energy-Consuming Architecture Evaluation}
\label{sec:evaluation}

If we perform the entire NAS only using the process described in \S\ref{sec:sampling}, \sys essentially is performing one-shot NAS within the sub-space $\mathcal{S}_{best}$. However, it is possible to improve LaMOO's space partition with more observed samples, as Zhao et al. showed~\cite{lamoo}. This section describes the process to evolve $\mathcal{S}_{best}$ during low carbon emission periods.

At the high level, we will pick new architectures to train to convergence and use them to refine the search space partition. That is, the architecture $\bm{a}$, with its true accuracy, $E(\bm{a})$, will be added to the observed sample set $\mathcal{P}$ to help identify a more advantageous sub-space, $\Omega_{best}$, for the architecture sampling process. In this work, we sort the architectures in the ready-to-train set $\mathcal{R}$ by their \emph{dominance number}. The dominance number $o({\bm{a}})$ of an architecture $\bm{a}$ is defined as the number of samples that dominate $\bm{a}$ in search space $\Omega$: 
\begin{equation}
o({\bm{a}}) := \sum_{{\bm{a}}_i \in \Omega}\mathbb{I}[{\bm{a}}_i \prec_f {\bm{a}},\ {\bm{a}} \neq {\bm{a}}_i],
\label{eq:dominance}
\end{equation}
where $\mathbb{I}[\cdot]$ is the indicator function. With the decreasing of the $o({\bm{a}})$, $\bm{a}$ would be approaching the Pareto frontier; $o(\bm{a})$ = $0$ when the sample architecture $\bm{a}$ locates in the Pareto frontier. The use of dominance number allows us to rank an architecture by considering both the estimated accuracy $E^{'}(\bm{a})$ and its inference energy cost $I(\bm{a})$ at the same time. \sys will first train the architectures with lower dominance number values when GPU resources are available. Once an architecture is trained, it is removed from $\mathcal{R}$. 

This process is depicted in the middle right part of Figure~\ref{fig:solution_overview}.
Note that this process includes actual time-consuming DL training, which is energy-intensive. Hence, \sys will try to prioritize this process during periods of low carbon intensity. 

\subsection{GPU Allocation Strategy}
\label{sec:allocation}

The carbon impact of the above two processes in a NAS search is materialized through the use of GPU resources. A key decision \sys needs to make is \emph{how} to allocate GPUs among these two interdependent processes. Assigning too many GPUs to the architecture sampling could impact the search efficiency, i.e., the searched architectures are far from the Pareto frontier; assigning too many GPUs to the architecture evaluation could significantly increase energy consumption. \sys must consider these trade-offs under varying carbon intensity and re-evaluate the GPU allocation strategy. 

Below we describe a heuristic strategy that automatically allocates GPU resources between the sampling and evaluation processes given the carbon emissions $C_{t}$ at time $t$. This allocation is based on the energy characteristics of the processes: architecture sampling is often energy-efficient because it does not involve actual training of architectures, while architecture evaluation is often energy-consuming because it does. 
We assume that the maximum and minimum carbon intensities $C_{max}$ and $C_{min}$ for a future time window are known. $G_{t}$ denotes the total number of available GPUs. $\lambda_{e}$ and $\lambda_{s}$ represent the ratio of GPU numbers allocated to the evaluation and sampling processes at a given moment, and $\lambda_{e} + \lambda_{s} = 1$. We calculate $\lambda_{s}$ as $\frac{C_{cur} - C_{min}}{C_{max} - C_{min}}$, where $C_{cur}$ is the current carbon intensity. The GPU allocations for the sampling and evaluation processes are, therefore, $G_{t} * \lambda_{s}$ and $G_{t} * \lambda_{e}$. This simple heuristic allocation allows \sys to prioritize more energy-efficient sampling tasks during periods of higher carbon intensity, whereas, during low-carbon periods, \sys will allocate more resources for energy-intensive evaluation tasks.

\section{Preliminary Results}

We prototype the \sys framework described in \S\ref{sec:roadmaps}. This section presents a preliminary analysis of \sys for its carbon and search efficiency based on trace-driven simulations. Specifically, we evaluate LaMOO's performance in partitioning the search space for NAS on HW-NASBench~\cite{hwnasbench}.

HW-NASBench was selected due to its inclusion of information on our two search targets: inference energy and accuracy. To assess the search performance and carbon cost of \sys, we have \sys search for optimal architectures on HW-NASBench and compare the searched results to three different NAS search methods. \sys delivers the most effective search results within the same carbon budget.

\begin{figure}[t]
\centering
\includegraphics[width=0.9\linewidth]{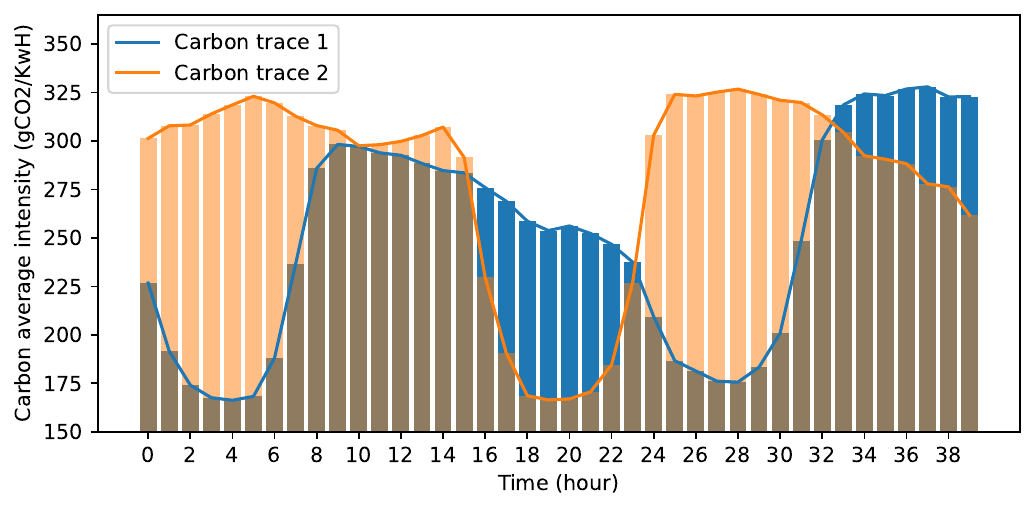}
\caption{Carbon traces from electricityMap.
\textnormal{
Trace 1 is based on the US-CAL-CISO data from 2021, specifically covering the period from 0:00, January 1, 2021, to 16:00, January 2, 2021. Trace 2 is also based on the US-CAL-CISO data from 2021, covering the period from 17:00, January 2, 2021, to 9:00, January 4, 2021.
}
}
\label{fig:trace}
\end{figure}

\begin{figure}[t]
\centering
\subfigure[Hypervolume]{\includegraphics[width=0.34\linewidth]{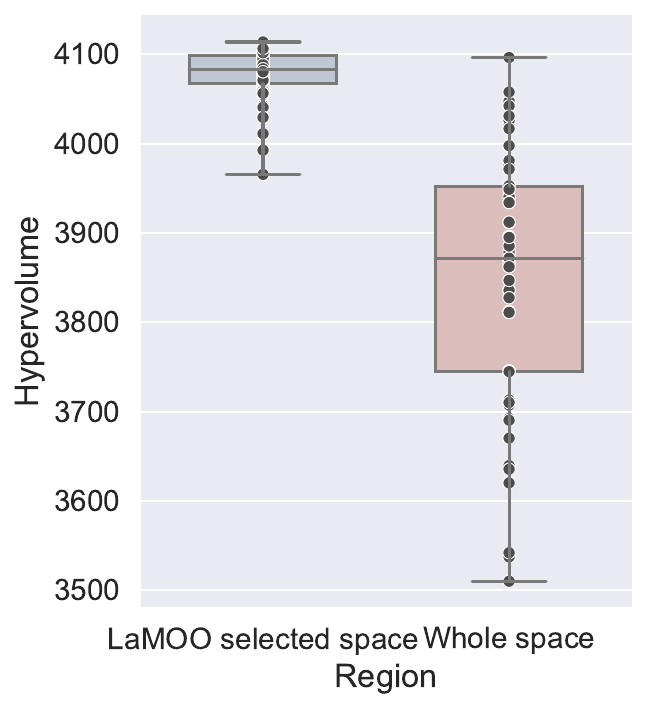}\label{fig:hwnasbench-c}}
\subfigure[Accuracy]{\includegraphics[width=0.32\linewidth]{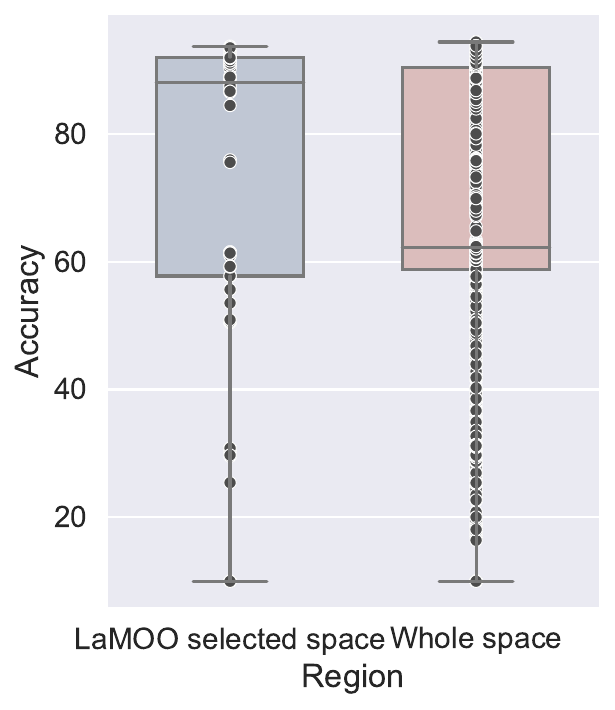}\label{fig:hwnasbench-a}}
\subfigure[Inference energy]{\includegraphics[width=0.32\linewidth]{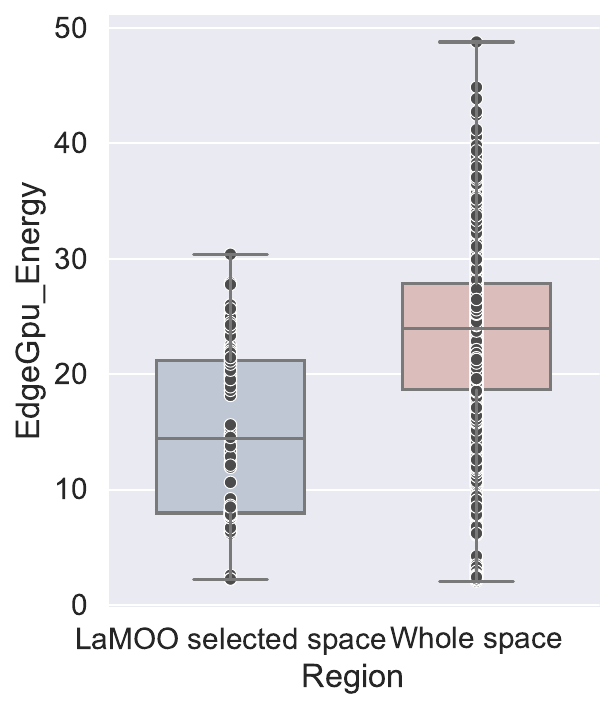}\label{fig:hwnasbench-b}}

\caption{
Comparisons of architecture qualities between LaMOO-selected region and the entire search space of HW-Nasbench.
\textnormal{
We ran LaMOO 10 times. For each run, we randomly sampled 50 architectures from the LaMOO-selected space and the whole search space. 
}
}
\label{fig:preliminary_hwnasbench}
\end{figure}

\begin{figure}[t]
\centering
\subfigure[With carbon trace 1]{\includegraphics[width=0.49\columnwidth]{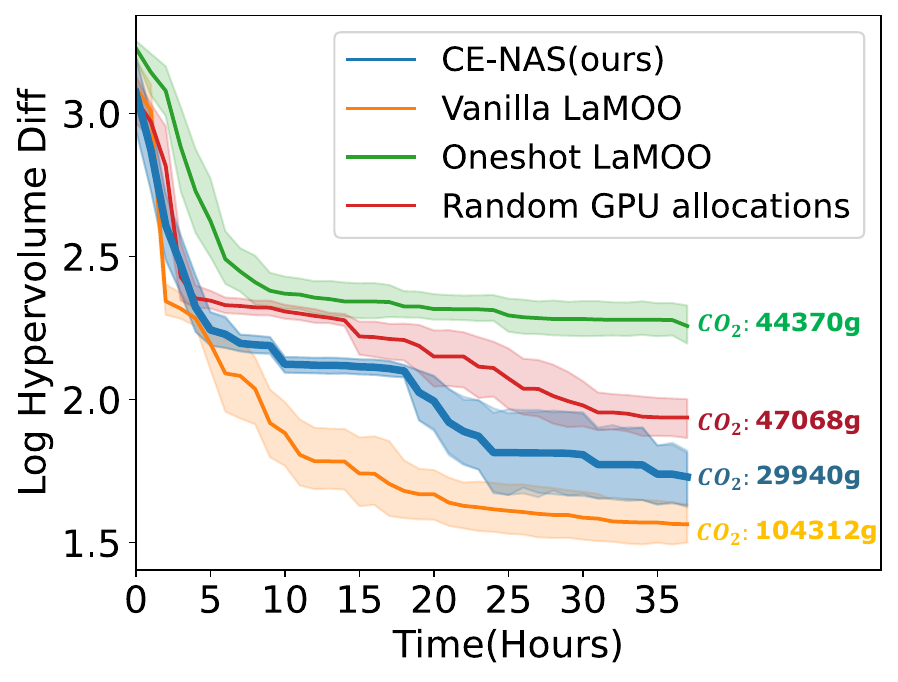}\label{fig:time-hv:trace1}}
\hfill
\subfigure[With carbon trace 2]{\includegraphics[width=0.49\columnwidth]{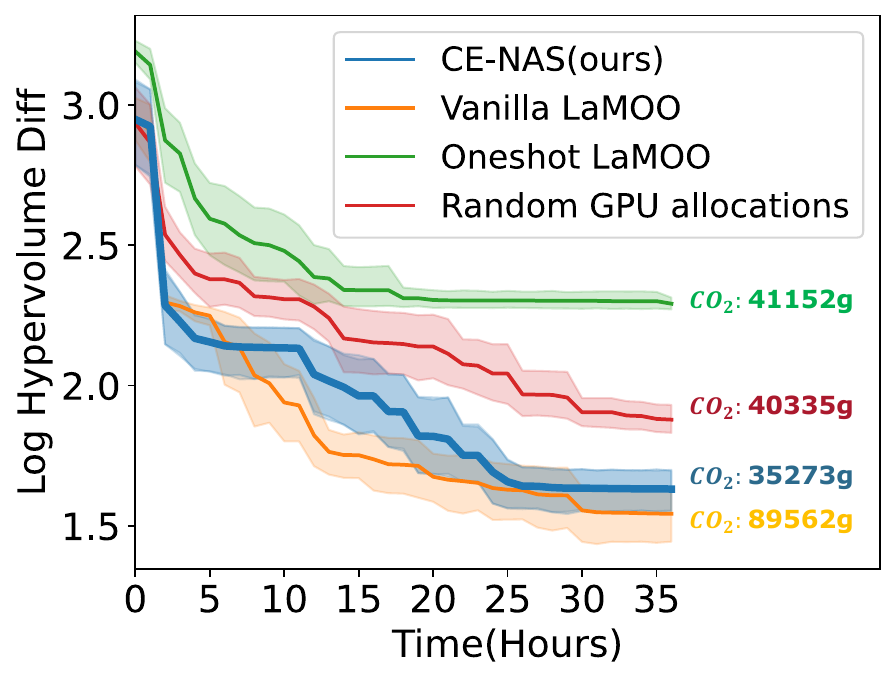}\label{fig:time-hv:trace2}}
\caption{Search progress over time. 
\textnormal{
\sys has the lowest relative carbon emission while achieving the second best $HV_{\mathrm{log\_diff}}$.
}
}
\label{fig:time-hv}
\end{figure}

\subsection{Setup}
\label{sec:setup}

We conduct our experiments using \sys and other baselines based on the two carbon traces depicted in Figure~\ref{fig:trace}. We initiate the process with ten samples in the set $\mathcal{P}$ and set the maximum capacity of $\mathcal{R}$ to be 300. Each method is simulated ten times for consistency, and all search processes in the simulation are executed on an Nvidia GeForce RTX 3090.

\para{Carbon Traces.} We used two carbon traces obtained from ElectricityMap~\cite{electricitymap}, a third-party carbon information service. Both carbon traces span 40 hours and consist of the per-hour average carbon intensity. 
We chose these two traces because they exhibit varying carbon intensity, as shown in Figure~\ref{fig:trace}, which allowed us to evaluate both the search over time performance and \sys's adaptiveness to carbon intensity.

\para{NAS Dataset.} A number of popular open-source NAS datasets, such as NasBench101~\cite{nasbench101}, NasBench201~\cite{nasbench201}, and NasBench301~\cite{nasbench301} exist. However, none contain information on architecture inference energy, one of our search objectives. We chose the new NAS dataset called \emph{HW-NASBench}~\cite{hwnasbench} due to its inclusion of information on our two search targets: inference energy and accuracy. Specifically, HW-NASBench contains inference performance of all networks in the NasBench201's search space~\cite{nasbench201} on six hardware devices, including commercial edge devices. In short, we use a combination of architecture information, including inference accuracy, training time, evaluation time, and energy cost in the edge GPU obtained from HW-NASBench and NasBench201~\cite{nasbench201}.

\begin{figure}[t]
\centering
\subfigure[$CO_{2}$ cost: 5000g]{\includegraphics[width=0.45\columnwidth]{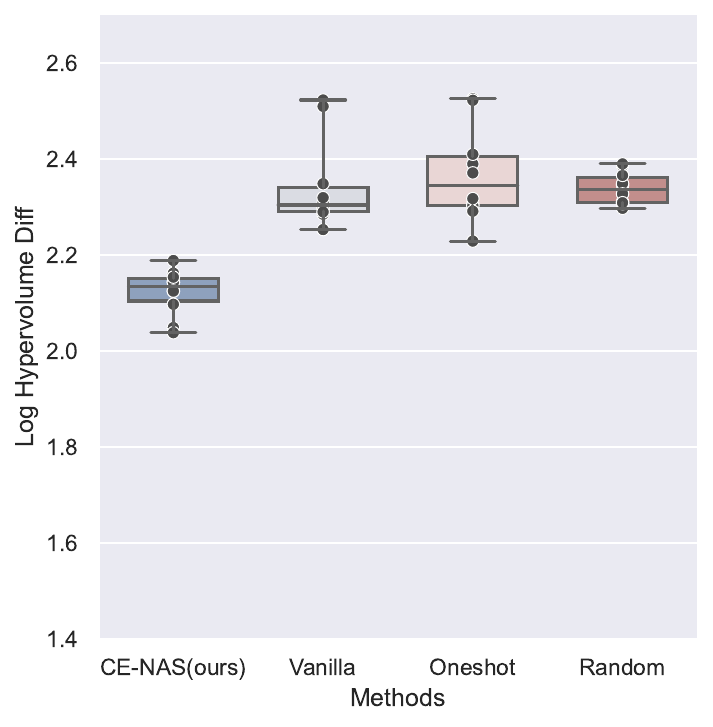}\label{fig:t0_5000}}
\hfill
\subfigure[$CO_{2}$ cost: 10000g]{\includegraphics[width=0.45\columnwidth]{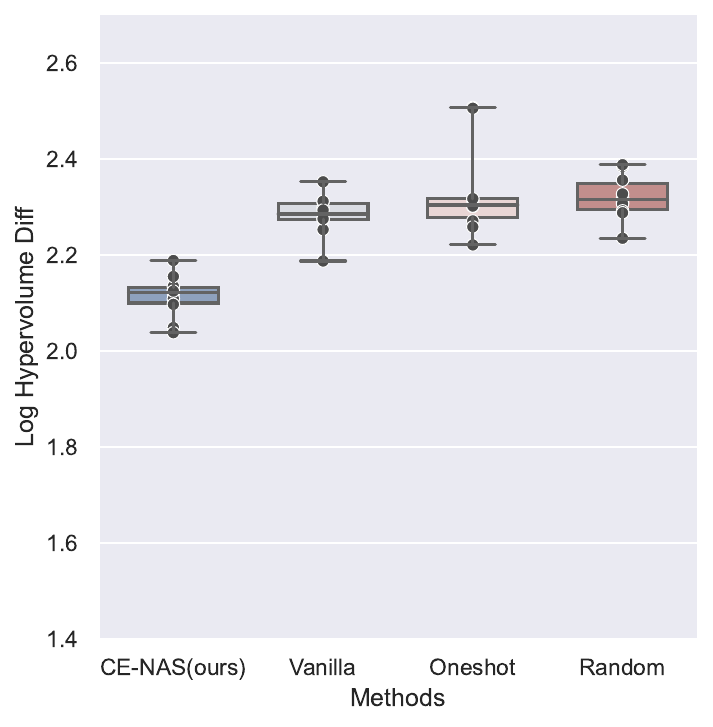}\label{fig:t0_10000}}
\hfill
\subfigure[$CO_{2}$ cost: 20000g]{\includegraphics[width=0.45\columnwidth]{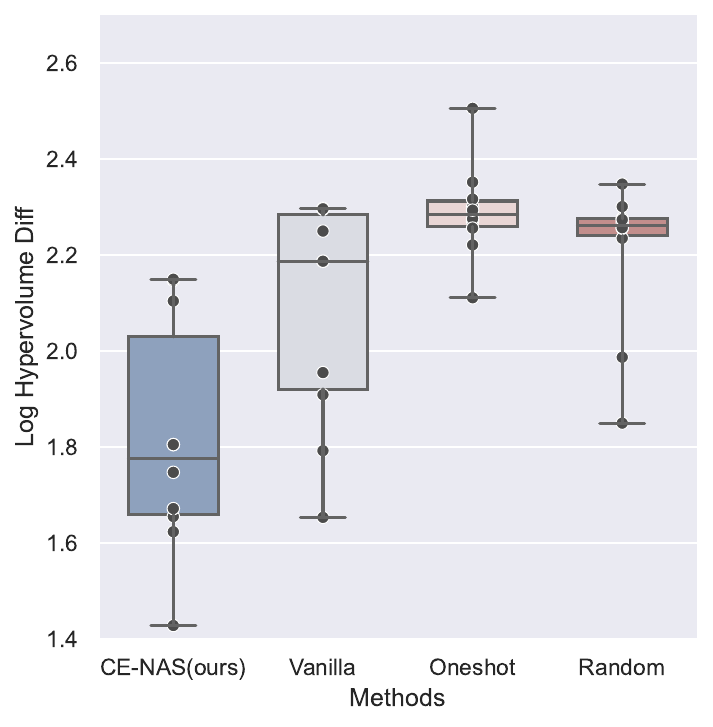}\label{fig:t0_20000}}
\hfill
\subfigure[$CO_{2}$ cost: 30000g]{\includegraphics[width=0.45\columnwidth]{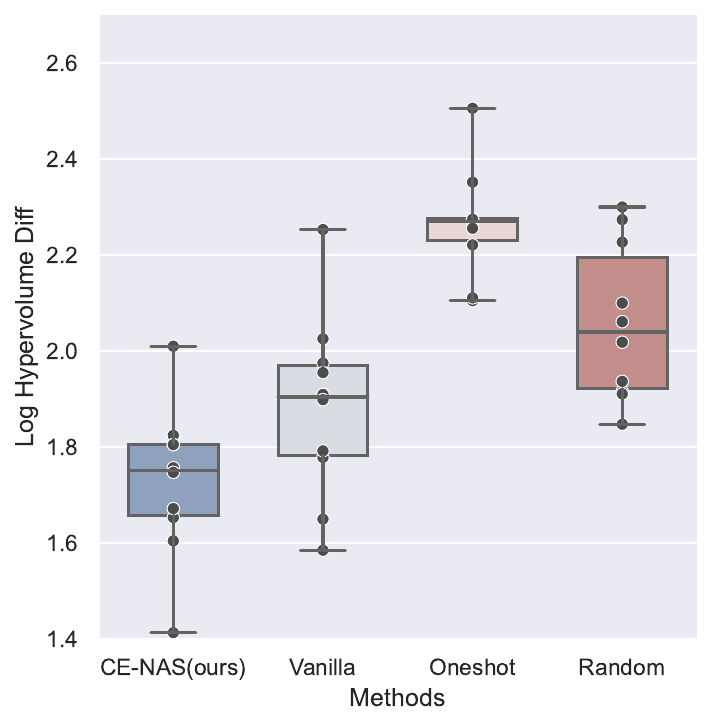}\label{fig:t0_30000}}
\caption{
Search efficiency under carbon emission constraints.
\textnormal{These results are obtained using carbon trace 1, and we ran each method ten times.}
}
\label{fig:barplot_t1}
\end{figure}

\begin{figure}[t]
\centering
\subfigure[$CO_{2}$ cost: 5000g]{\includegraphics[width=0.45\columnwidth]{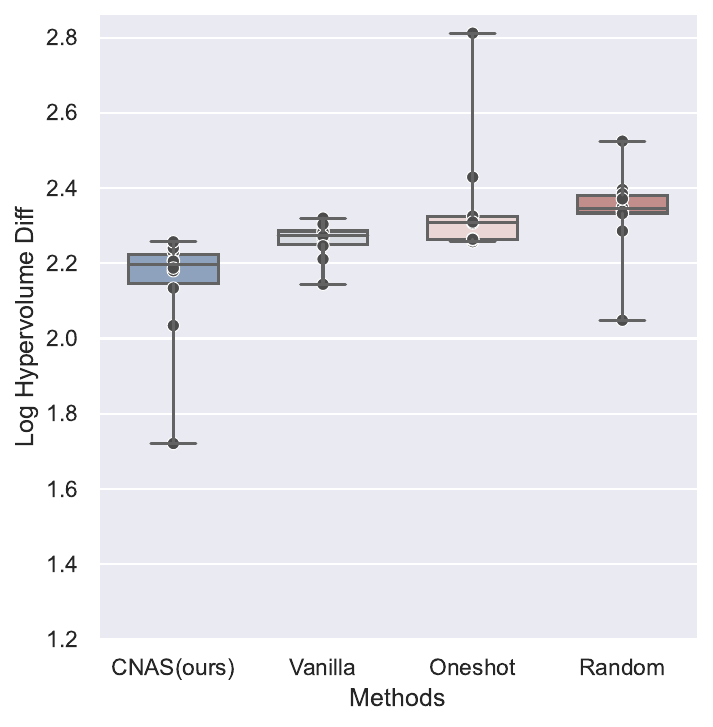}\label{fig:t0_5000}}
\hfill
\subfigure[$CO_{2}$ cost: 10000g]{\includegraphics[width=0.45\columnwidth]{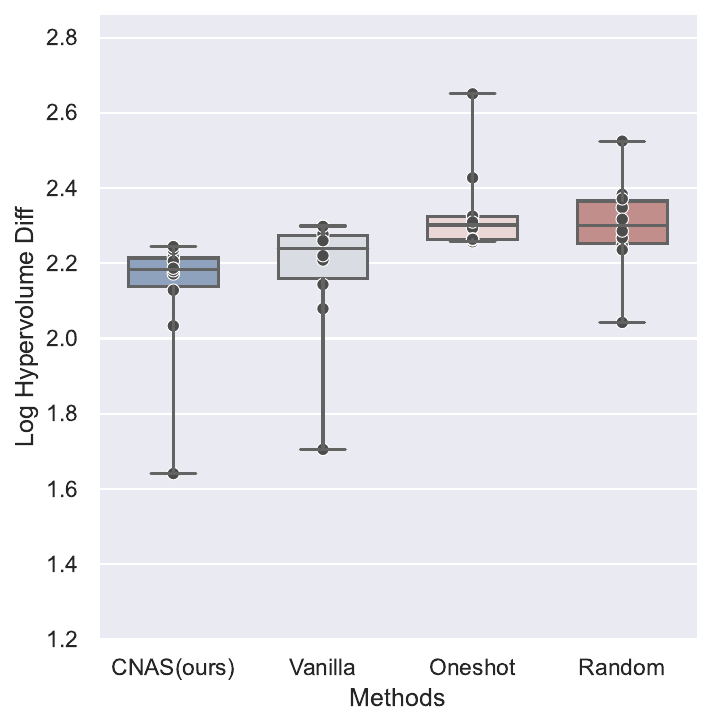}\label{fig:t0_10000}}
\hfill
\subfigure[$CO_{2}$ cost: 20000g]{\includegraphics[width=0.45\columnwidth]{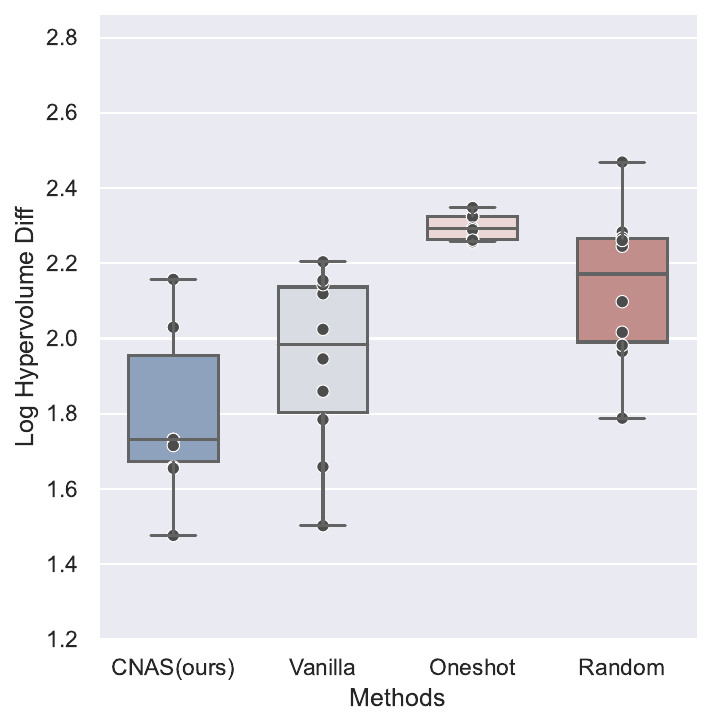}\label{fig:t0_20000}}
\hfill
\subfigure[$CO_{2}$ cost: 30000g]{\includegraphics[width=0.45\columnwidth]{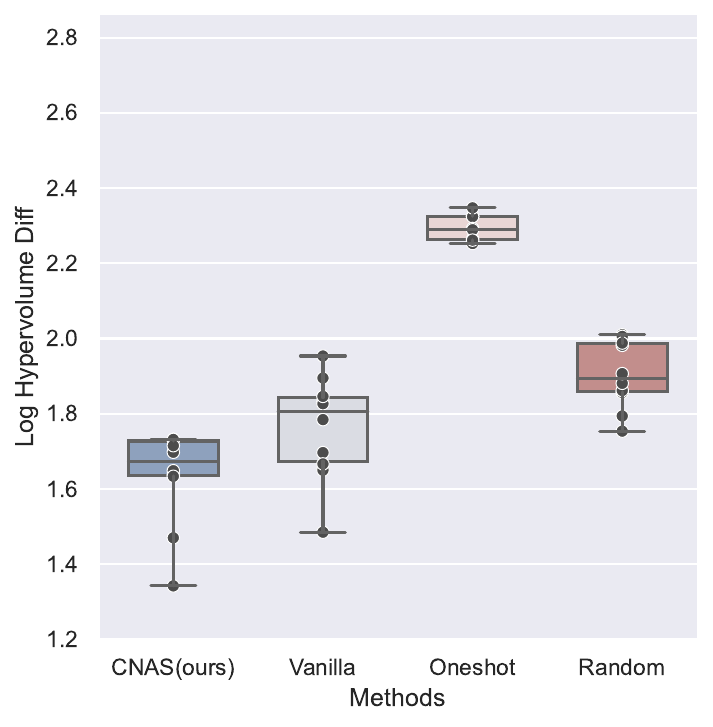}\label{fig:t0_30000}}
\caption{
Search efficiency under carbon emission constraints.
\textnormal{These results are obtained using carbon trace 2, and we ran each method ten times.}
}
\label{fig:barplot_t2}
\end{figure}

\para{Metrics.} We use two main metrics to evaluate the carbon and search efficiency of \sys. First, we use \emph{relative carbon emission} to quantify the amount of CO2 each NAS method is responsible for. The relative carbon emission is calculated by summing the average carbon intensity (in gCO2/KwH) over the search process. We assume that all NAS methods use the same type of GPU whose power consumption remains the same throughout the search process. Second, we use the metric \emph{hypervolume} (HV) to measure the "goodness" of searched samples. HV is a commonly used multi-objective optimization quality indicator~\cite{qehvi, qnehvi, lamoo} that considers all dimensions of the search objective. Given a reference point $R \in \rr^M$, the HV of a finite approximate Pareto set $\mathcal{P}$ is the M-dimensional Lebesgue measure $\lambda_{M}$ of the space dominated by $\mathcal{P}$ and bounded from below by $R$. That is, $HV(\mathcal{P}, R) = \lambda_{M} (\cup_{i=1}^{|\mathcal{P}|}[R, y_{i}])$, where $[R, y_{i}]$ denotes the hyper-rectangle bounded by the reference point $R$ and $y_{i}$. A higher hypervolume denotes better multi-objective results.

\para{Baselines.} 
We chose three types of baselines according to different GPU allocation strategies and NAS evaluation algorithms. During the search process, all search methods employ the state-of-the-art multi-objective optimizer, LaMOO~\cite{lamoo}. 
Specifically, \emph{one-shot LaMOO} is a method that utilizes one-shot evaluations throughout the search process. The \emph{vanilla LaMOO} relies on actual training for architecture evaluation throughout the search. The \emph{random GPU allocations} is a strawman strategy that randomly allocates GPUs between the energy-efficient sampling stage and the more energy-consuming evaluation stage without considering the carbon intensity.

\subsection{Effectiveness of LaMOO for NAS}

We conducted ten runs of LaMOO (i.e., search space split) with a random search on the HW-NASBench dataset~\cite{hwnasbench}. In addition, we performed random sampling for both the LaMOO-selected region and the entire search space, conducting 50 trials for each. 
The distribution of accuracy and edge GPU energy consumption of the architectures in both the LaMOO selected region and the entire search space can be seen in Figure~\ref{fig:preliminary_hwnasbench}. 

Specifically, our results show that the architectures in the region selected by LaMOO have higher average accuracy and lower average edge GPU energy consumption compared to those in the entire search space. On average, the accuracy of the architectures in the LaMOO selected region is 72.12, while the accuracy in the entire search space is 68.28. The average edge GPU energy for the LaMOO selected region is 16.59 mJ, as opposed to 22.84 mJ for the entire space.

Furthermore, as illustrated in Figure~\ref{fig:hwnasbench-c}, we observe that searching within the LaMOO-selected region yielded a tighter distribution, and the median hypervolume demonstrated an improvement compared to searching across the entire search space.
These results suggest the efficacy of using LaMOO to partition the search space for NAS.

\subsection{Carbon and Search Efficiency}
\label{subsec:carbon_search_efficiency}

In this section, we evaluate the search performance and carbon costs of our \sys framework, comparing it to three other baselines on the HW-NASBench dataset~\cite{hwnasbench}. We use the \emph{log hypervolume} difference, the same as in~\cite{qehvi, qnehvi, lamoo}, as our evaluation criterion for HW-NASBench, since the hypervolume difference may be minimal over the search process. Therefore, using log hypervolume allows us to visualize the sample efficiency of different search methods. 
We define $HV_{\mathrm{log\_diff}} := \log(HV_{\mathrm{max}} - HV_{\mathrm{cur}})$ 
where $HV_{\mathrm{max}}$ represents the maximum hypervolume calculated from all points in the search space, and  $HV_{\mathrm{cur}}$ denotes the hypervolume of the current samples, which are obtained by the algorithm within a specified budget. The $HV_{\mathrm{max}}$ in this problem is 4150.7236.  For our simulation, we use the training and evaluation time costs for the architectures derived from NasBench201~\cite{nasbench201}, and inference energy costs measured on the NVIDIA Edge GPU Jetson TX2 from HW-NASBench~\cite{hwnasbench}. 
We ran the simulation 10 times with each method.

As depicted in Figure~\ref{fig:time-hv}, as the search time progresses, vanilla LaMOO demonstrates the highest performance in terms of $HV_{\mathrm{log\_diff}}$. Vanilla LaMOO's superior performance can be attributed to its approach of \emph{training all sampled architectures} to obtain their true accuracy, which effectively steers the search direction. However, when considering the relative carbon emission, vanilla LaMOO consumes 2.22X-3.48X carbon compared to other approaches. This is expected because vanilla LaMOO is an energy-consuming approach and is not designed to be aware of carbon emissions associated with joules.

We show that \sys's search efficiency is only marginally lower than that of vanilla LaMOO while having the least relative carbon emission under both carbon traces. 
Note that we are plotting the $HV_{\mathrm{log\_diff}}$ in the Y-axis of Figure~\ref{fig:time-hv}; the actual $HV$ values achieved by \sys and Vanilla LaMOO are about 4100 and 4117, differing only by 0.034\%, even though the two lines look far apart. 
This result suggests that only relying on energy-efficient approaches (e.g., one-shot LaMOO in this case) is insufficient to achieve carbon efficiency. For both traces, one-shot LaMOO has 1.17X-1.48X carbon compared to \sys.

Moreover, we observe that \sys's carbon efficiency is correlated to the time-varying carbon intensity. When the coefficient of variation of carbon intensity is higher, \sys has more opportunity to explore the GPU allocation trade-offs between energy-efficient sampling and energy-consuming evaluation without impacting search quality. The relative carbon emission difference between \sys and \emph{random GPU allocations} represents how well \sys makes such trade-offs. Currently, we are using a heuristic approach, and it is possible to devise more sophisticated strategies to further reduce relative carbon emissions. 
For example, if the strategy could determine the GPU resources based on the queued architectures and the current carbon intensity, it can better shift the workload to periods of low carbon emission.

Finally, Figure~\ref{fig:barplot_t1} and ~\ref{fig:barplot_t2} compare \sys performance with baselines under different carbon budgets. We show that \sys outperforms all baselines in terms of search efficiency. This is because when there is a carbon budget, energy-consuming approaches (e.g., vanilla LaMOO) would exhaust the budget and end the search earlier, as opposed to operating with an unlimited carbon budget. This result suggests \sys's ability to dynamically adjust the search process based on carbon budgets while still producing reasonable search efficiency.

\section{Conclusion and Future Directions}
\label{sec:conclusion}

In this work, we described the design of a carbon-efficient NAS framework \sys by leveraging the temporal variations in carbon intensity. To search for energy-efficient architectures, \sys integrates a state-of-the-art multi-objective optimizer, LaMOO~\cite{lamoo}, with the one-shot and vanilla NAS algorithms. These two NAS evaluation strategies have different energy requirements, which \sys leverages to schedule when to use each based on average carbon intensity. Our trace-driven simulations show that \sys is a promising approach for reducing relative carbon emission while maintaining search efficiency. 

Based on our investigation, we believe there are many fruitful directions in the context of \sys. For example, one can train an agent, e.g., use deep reinforcement learning, to automatically output different GPU allocation strategies based on historical carbon traces. This can replace our current heuristic GPU allocation strategy and will likely lead to better carbon and search efficiency. Another direction is to develop models that are capable of accurately predicting carbon intensity, similar to a recent work~\cite{carboncast}. With such predictive models, \sys can better schedule the NAS tasks to a dynamic set of GPUs that can span across geographic locations without adversely impacting the total search time.

\begin{acks}
This work was supported in part by NSF Grants \#2105564 and \#2236987, and a VMWare grant. We also thank electricityMap for its carbon intensity dataset.
\end{acks}

\balance
\bibliographystyle{ACM-Reference-Format}
\bibliography{ce-nas}

\end{document}